\documentclass[sigconf]{acmart}
\AtBeginDocument{%
  }

\usepackage{multirow}
\usepackage{amsfonts}
\usepackage{subfig}

\newtheorem{proposition}{\textbf{Proposition}}[section]

\newtheorem{theorem}{\textbf{Theorem}}[section]
\copyrightyear{2026}
\acmYear{2026}
\setcopyright{othergov}
\acmConference[MM '26]{Proceedings of the 34th ACM International Conference on Multimedia}{November 10--14, 2026}{Rio de Janeiro, Brazil}
\acmBooktitle{Proceedings of the 34th ACM International Conference on Multimedia (MM '26), November 10--14, 2026, Rio de Janeiro, Brazil}
\acmDOI{10.1145/3767308.3835335}
\acmISBN{979-8-4007-2213-4/2026/11}

\acmSubmissionID{2471}

\makeatletter
\gdef\@copyrightpermission{
\begin{minipage}{0.3\columnwidth}
 \href{https://creativecommons.org/licenses/by/4.0/}{\includegraphics[width=0.90\textwidth]{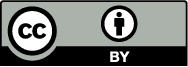}}
  \end{minipage}\hfill
\begin{minipage}{0.7\columnwidth}
 
 \href{https://creativecommons.org/licenses/by/4.0/}{This work is licensed under a Creative Commons Attribution International 4.0 License.}
 \end{minipage}
 \vspace{5pt}
}
\makeatother

\begin{document}

%%
%% The "title" command has an optional parameter,
%% allowing the author to define a "short title" to be used in page headers.
\title{Closing the Loop: A Control-Theoretic Framework for Time Series Forecasting with LLMs}

%%
%% The "author" command and its associated commands are used to define
%% the authors and their affiliations.
%% Of note is the shared affiliation of the first two authors, and the
%% "authornote" and "authornotemark" commands
%% used to denote shared contribution to the research.
\author{Xingyu~Zhang}
\orcid{0009-0001-4216-9870}
% \authornotemark[1]
\authornote{These authors have contributed equally}
 \affiliation{
  \institution{Institute of Software Chinese Academy of Sciences}
\city{Beijing}
  \country{China}
 }  
\affiliation{
  \institution{University of Chinese Academy of Sciences}
    \city{Beijing}
  \country{China}
}
\email{zhangxingyu23@mails.ucas.ac.cn}

\author{Jingyao~Wang}
\orcid{0000-0003-1782-8704}
\authornotemark[1]
% \authornote{These authors have contributed equally}
 \affiliation{
  \institution{Institute of Software Chinese Academy of Sciences}
\city{Beijing}
  \country{China}
 }  
\affiliation{
  \institution{University of Chinese Academy of Sciences}
    \city{Beijing}
  \country{China}
}
\email{wangjingyao2023@iscas.ac.cn}

\author{Zeen~Song}
\orcid{0000-0003-4064-4129}
\affiliation{%
  \department{Digital-Intelligent Operations Center}
  \institution{Guizhou Power Grid Co., Ltd.}
  \city{Guizhou}
  \country{China}}
\email{zeensong@qq.com}

\author{Changwen~Zheng}
\orcid{0000-0002-2311-6757}
\affiliation{
  \institution{Institute of Software Chinese Academy of Sciences}
\city{Beijing}
  \country{China}
 }  
\email{changwen@iscas.ac.cn}

\author{Wenwen~Qiang}
\orcid{0000-0002-7985-5743}
\authornote{Corresponding Author}
% \affiliation{%
%   \institution{Institute of Software Chinese Academy of Sciences}
%   % \streetaddress{1 Th{\o}rv{\"a}ld Circle}
% }  
\affiliation{
  \institution{Institute of Software Chinese Academy of Sciences}
\city{Beijing}
  \country{China}
 }  
\email{qiangwenwen@iscas.ac.cn}

%%
%% By default, the full list of authors will be used in the page
%% headers. Often, this list is too long, and will overlap
%% other information printed in the page headers. This command allows
%% the author to define a more concise list
%% of authors' names for this purpose.
\renewcommand{\shortauthors}{Xingyu Zhang, Jingyao Wang, Zeen Song, Changwen Zheng, and Wenwen Qiang}
%%
%% The abstract is a short summary of the work to be presented in the
%% article.
\begin{abstract}
Large Language Models (LLMs) have recently shown exceptional potential in time series forecasting (TSF), leveraging their inherent sequential reasoning capabilities to model complex temporal dynamics. Existing approaches typically employ an autoregressive generation strategy to adapt LLMs for TSF. However, we identify a theoretical flaw in this paradigm: during inference, the model operates in an open-loop manner, recursively consuming its own generated outputs. This leads to error accumulation, where minor early deviations cascade into significant rollout drift over long horizons. In this paper, we reformulate autoregressive forecasting through the lens of control theory, proposing Feedback-driven LLM (F-LLM), a novel closed-loop framework. Unlike standard methods that passively propagate errors, F-LLM actively stabilizes the trajectory via a learnable residual estimator functioning as a system observer. Furthermore, we provide a mathematical proof that, under explicit contraction assumptions, this closed-loop mechanism guarantees a uniformly bounded step-wise error sequence within the local surrogate dynamics. Extensive experiments demonstrate that F-LLM significantly mitigates error propagation, achieving good performance on time series benchmarks. Our code is publicly available at \url{https://github.com/Zh-XY22/F-LLM}.
\end{abstract}

%%
%% The code below is generated by the tool at http://dl.acm.org/ccs.cfm.
%% Please copy and paste the code instead of the example below.
%%


%%
%% Keywords. The author(s) should pick words that accurately describe
%% the work being presented. Separate the keywords with commas.
%% A "teaser" image appears between the author and affiliation
%% information and the body of the document, and typically spans the
%% page.

%%
%% This command processes the author and affiliation and title
%% information and builds the first part of the formatted document.
\maketitle

\begin{figure*}
    \centering
    \includegraphics[width=0.82\linewidth]{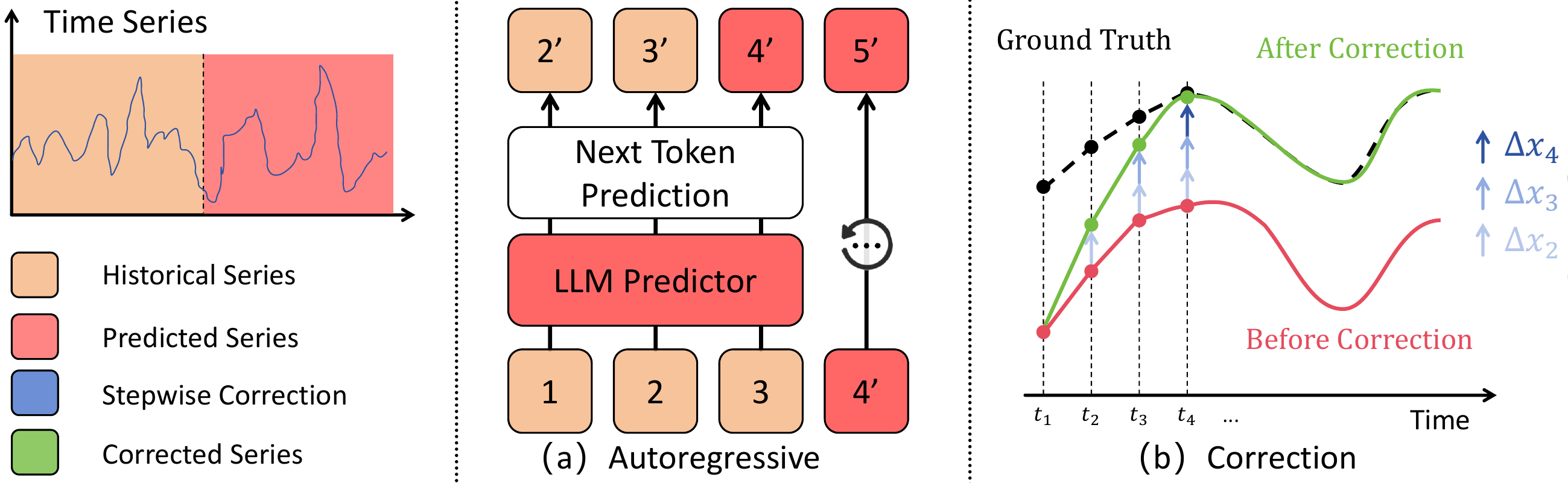}
    \caption{Comparison of standard autoregressive generation and the proposed stepwise correction strategy. (a) Standard autoregressive LLMs suffer from cumulative prediction drift, where minor deviations at each step are fed back as input, causing errors to accumulate over time. (b) The stepwise correction strategy intervenes in this process by adjusting the prediction at each step before it enters the next autoregressive loop, thereby preventing error propagation and stabilizing the trajectory.}
    \label{fig:intro_llm_bias}
\end{figure*}

\section{Introduction}
% [Paragraph 1: 背景]
Time Series Forecasting (TSF) stands as a pivotal component in modeling continuous sensor streams \cite{MM23TSF, MM25Foresail}, underpinning critical spatial-temporal applications ranging from energy grid monitoring \cite{TSF2ECL, TSF2ECL2} to extreme weather event prediction \cite{TSF2weather, TSF2weather2}. Recently, the adaptation of large language models (LLMs) for time series analysis has attracted growing interest \cite{FPT}. Recent studies demonstrate that aligning continuous forecasting tasks with the autoregressive generation paradigm of LLMs can effectively unleash their immense pre-trained potential \cite{Time-LLM, Lag-Llama}, as shown in Fig.\ref{fig:intro_llm_bias}(a). This alignment allows models to preserve temporal coherence, capture long-range dependencies, and exhibit remarkable zero-shot generalization.

% [Paragraph 2: 提出问题 + 解释]
However, directly applying the discrete next-token prediction paradigm to continuous time series introduces a structural vulnerability: error accumulation. In natural language processing, minor token variations are often semantically tolerated. In continuous domains, however, current states determine future trajectories. This inherent sensitivity exacerbates a fundamental mismatch known as exposure bias \cite{bengio2015scheduled}. Specifically, standard LLMs are trained using ground-truth history as context, but during inference, they operate in an open-loop autoregressive mode, recursively consuming their own generated outputs. Consequently, even a microscopic prediction error at one step alters the input context for the next. Without ground-truth correction as Fig.\ref{fig:intro_llm_bias}(b), these errors propagate and amplify through the autoregressive loop \cite{venkatraman2015improving}. This ``snowball effect'' causes the predicted sequence to deviate significantly from the actual future path, making naive autoregressive LLMs susceptible to severe rollout drift as the forecasting horizon extends. Thus, in this paper, we mainly focus on: \emph{How to mitigate the error accumulation when using LLM for TSF?}

% [Paragraph 3: 问题剖析 / Motivation 实验]
To answer this, we theoretically analyze the error propagation mechanism of autoregressive forecasting in Section \ref{sec:theory}. We demonstrate that under locally expansive rollout dynamics, the upper bound of open-loop prediction errors amplifies geometrically. Because the autoregressive transition acts as a recursive multiplier, even bounded single-step noise compounds exponentially over long horizons. This reveals a critical bottleneck: simply reducing single-step prediction error is mathematically insufficient to prevent long-term rollout drift. Consequently, mitigating this drift requires a shift in the generation dynamics.

% [Paragraph 4: 具体 Motivation (回应剖析)]
To intervene at the system level, inspired by feedback control theory, we propose reformulating autoregressive inference from an open-loop sequence generation into a closed-loop dynamic process. 
Rather than passively accepting accumulated errors during rollout, this closed-loop paradigm allows us to actively estimate the latent error state and compensate for it on-the-fly. By dynamically calibrating the generation trajectory, we can maintain the predictions within a bounded deviation from the desired rollout.

% [Paragraph 5: 具体方法 + 实验结果]
Guided by these theoretical insights, we propose \textbf{F-LLM}, named Feedback-driven LLM, to instantiate this closed-loop system. As demonstrated in Section \ref{sec:theory}, practical stabilization requires addressing two critical conditions. First, since the true prediction error is unobservable during inference, F-LLM introduces a lightweight residual estimator functioning as a system observer to infer the latent error and construct the feedback signal. Second, motivated by the stability conditions of the closed-loop system, we impose a Local Lipschitz constraint on the projection layers to reduce excessive local sensitivity, making the rollout dynamics amenable to feedback correction. Extensive experiments on standard benchmarks show that our closed-loop intervention mitigates rollout drift, achieving consistent improvements in long-horizon forecasting accuracy.
% [Paragraph 6: Contribution]
The main contribution can be summarized as: 
\begin{itemize}
    \item We provide an analysis of the error propagation mechanism in autoregressive LLMs, demonstrating that open-loop inference can suffer from geometric error amplification under locally expansive dynamics.
    \item We propose F-LLM, a novel feedback-driven framework that integrates deep generative models with closed-loop control to mitigate rollout drift in LLM-based forecasting.
    \item We provide a mathematical analysis showing that, under explicit boundedness and contraction assumptions, the closed-loop mechanism yields a uniformly bounded step-wise error sequence in the local surrogate dynamics.
    \item Extensive experiments on various benchmark datasets demonstrate that F-LLM mitigates rollout drift, achieving superior performance in long-term forecasting tasks.
\end{itemize}

\section{Related Works}

\subsection{Time Series Forecasting}
With the advent of deep learning \cite{lecun_deep_2015}, the field of TSF has witnessed a transition from statistical methods to data-driven neural architectures. These approaches can be broadly categorized based on their core mechanisms: RNN-based models \cite{how_RNN2TSF, tang2021building} capture sequential dependencies; CNN-based methods \cite{zhan2023differential, bai2018empirical} extract local temporal features; and MLP-based architectures \cite{zhang2024not, DLinear} emphasize efficient point-wise mapping. Recently, Transformer-based models have dominated the landscape, focusing on different modeling perspectives: temporal-domain methods like PatchTST \cite{PatchTST} and TimesNet \cite{TimesNet} capture multi-scale dependencies; frequency-domain approaches such as FEDFormer \cite{FEDformer} utilize Fourier transforms for global modeling; and cross-variable methods like iTransformer \cite{iTransformer} explicitly model multivariate correlations. 
% However, these models operate primarily as discriminative regressors, typically requiring extensive training on domain-specific data, which limits their zero-shot generalization capabilities across diverse scenarios. 
Beyond traditional domains, TSF is increasingly pivotal in multimedia systems, such as predicting video streaming traffic, forecasting multimedia content popularity, and analyzing continuous multimodal sensor streams.

\subsection{LLMs for TSF}
Recognizing the generalization power of foundation models, recent research explores adapting LLMs for TSF through two distinct paradigms. The first stream treats the LLM primarily as a deep feature extractor followed by a global projection head. Methods such as FPT \cite{FPT}, Time-LLM \cite{Time-LLM}, TEST \cite{TEST} and PromptCast \cite{PromptCast}, reprogram the input time series into text-aligned prototypes or embeddings, but eventually rely on a linear layer to predict the entire future horizon in a single forward pass. While this avoids error accumulation, it fundamentally reduces the generative LLM to a discriminative regressor, failing to fully exploit the model's intrinsic capacity for sequential reasoning and probabilistic modeling. To better align with the LLM's pre-training objective, a second stream advocates for autoregressive forecasting, represented by LLMTime \cite{LLMTime}, Chronos \cite{Chronos}, Lag-Llama \cite{Lag-Llama}, and AutoTimes \cite{AutoTimes}. By staying faithful to the ``next-token'' paradigm, these methods model the joint probability of the future sequence to capture complex temporal dependencies. However, a critical limitation persists across these autoregressive generation methods: they rely on a naive, open-loop generation process. By recursively consuming their own uncorrected outputs, these methods remain highly susceptible to error accumulation, rendering long-horizon forecasts prone to severe rollout drift. Unlike these open-loop paradigms, our work fundamentally reshapes the autoregressive generation process. By introducing a closed-loop feedback mechanism, we actively estimate and compensate for step-wise errors during inference, thereby preventing the ``snowball effect'' of error accumulation.

\subsection{Exposure Bias}
The discrepancy between training with ground truth context and generating sequences autonomously during inference, known as exposure bias, is a classical challenge in sequence generation \cite{bengio2015scheduled}. Traditional mitigation strategies, such as Scheduled Sampling \cite{bengio2015scheduled} or sequence-level optimization via reinforcement learning \cite{ExposureBias, wiseman2016sequence}, require intervening in the training process. Recent multimedia generation studies have also explored richer reward modeling and Pareto-aware multi-reward alignment \cite{Ba_2025_ICCV,ba2026paretoguidedoptimaltransportmultireward}, although their focus is preference optimization rather than autoregressive stability. For LLM-based forecasting, however, retraining is often computationally prohibitive or infeasible due to restricted access to model weights. Consequently, inference-time solutions are preferred. While simple heuristic adjustments exist \cite{li2022contrastive, chuang2024dola}, they fail to address the fundamental dynamics of error accumulation in continuous time series. This highlights a significant gap in the literature: in the context of TSF, there remains a lack of a rigorous, control-theoretic framework that can stabilize the autoregressive generation of frozen LLMs without necessitating expensive retraining \cite{leaf, KAIROS}. Our proposed F-LLM directly employs a residual estimator to construct an explicit error-correction feedback loop and imposes Local Lipschitz regularization to bound system sensitivity. We provide a theoretically grounded, lightweight solution to stabilize long-horizon rollouts.

\section{Notation and Problem Formulation}
We begin by formalizing the task of TSF under the autoregressive paradigm of LLMs, and then highlight the theoretical mismatch that arises when this paradigm is directly applied to TSF.

\subsection{TSF under the Autoregressive Paradigm}
Let $x_{1:T} \in \mathbb{R}^{T \times D}$ denote a multivariate historical time series of length $T$ with $D$ variables, and let the forecasting target be the future horizon $x_{T+1:T+H} \in \mathbb{R}^{H \times D}$. Traditional forecasting methods typically formulate this problem as a direct mapping $f: \mathbb{R}^{T \times D} \rightarrow \mathbb{R}^{H \times D}$. In contrast, LLM-based approaches adopt a generative formulation. Specifically, an LLM parameterized by $\theta$ acts as a probabilistic predictor $g_\theta$ that models the joint distribution of the future sequence through autoregressive factorization. By the chain rule, this model can be written as
\begin{equation}\label{qweqwe}
% \resizebox{0.9\linewidth}{!}{$
P(x_{T+1:T+H} \mid x_{1:T}) = \prod_{k=1}^{H} P(x_{T+k} \mid x_{1:T+k-1}).
% $}
\end{equation}
This formulation is consistent with the next-token prediction objective commonly used in LLM pre-training, thereby enabling the model to exploit its pre-trained priors for sequential modeling and zero-shot generalization~\cite{AutoTimes}.

\subsection{Open-Loop Vulnerability in Forecasting}
Despite its powerful generative capabilities, the autoregressive inference process introduces an inherent structural vulnerability, namely error accumulation. During forecasting, the model operates in an open-loop manner, where the prediction at step $T+k$ depends on previously generated estimates rather than ground-truth observations. Then, the recursive generation process is given by
\begin{equation}
\hat{x}_{T+k} = g_\theta(\hat{X}_{T+k-1}),
\end{equation}
where $\hat{X}_{T+k-1}$ denotes the current context consisting of the historical input together with the previously generated outputs. As a result, the model is trained under a ground-truth data distribution but deployed under an autoregressive generation distribution. This mismatch is widely known as exposure bias in sequence generation~\cite{bengio2015scheduled,ExposureBias}. Small prediction errors made at early steps may propagate and amplify through the feedback loop \cite{ross2011reduction}. In continuous dynamical settings, this effect often leads to compounding error growth over the forecasting horizon~\cite{venkatraman2015improving}. Therefore, robust long-horizon forecasting cannot rely solely on recursive generation, but instead requires reformulating inference as a stabilized dynamical or control process equipped with explicit error correction.

\section{Theoretical Analysis}
\label{sec:theory}

In this section, we analyze why autoregressive forecasting may become unstable over long horizons and show how residual feedback changes the error dynamics. Our goal is not to claim that every open-loop forecaster necessarily diverges, but to identify a practically relevant regime in which locally expansive rollout dynamics lead to geometric error amplification, and to derive a stability condition that motivates our design.

\subsection{Open-Loop Error Propagation}
For theoretical analysis, we introduce an effective autoregressive state $s_t$ that summarizes the historical context available at step $t$, the predictor produces the next-step forecast from this state. Let $\hat{x}_t$ and $x_t$ denote the predicted and ground-truth target at generation step $t$, respectively. Define the prediction error as
\begin{equation}
    \Delta_t = x_t - \hat{x}_t .
\end{equation}
Under a first-order local approximation around a reference rollout, the open-loop error dynamics can be written as
\begin{equation}
    \Delta_{t+1} = A_t \Delta_t + \xi_t,
    \label{eq:open_loop_general}
\end{equation}
where $A_t \in \mathbb{R}^{D \times D}$ describes how the current prediction error affects the next-step error. In other words, it characterizes the local sensitivity of the autoregressive rollout to small perturbations around the current trajectory. The term $\xi_t$ includes one-step prediction errors and non-linear residuals. Eq.~(\ref{eq:open_loop_general}) shows that long-horizon forecasting performance depends not only on the current prediction error itself, but also on whether this error is amplified or suppressed during rollout. If the autoregressive transition is locally non-contractive, then even a small early error may be propagated and enlarged over time, resulting in accumulated deviation over the forecasting horizon. This observation is formalized below.

\begin{proposition}[Upper Bound for Open-Loop Error]
\label{prop:growth}
Assume that over a rollout of length $H$, the local sensitivity satisfies $\|A_{t+i}\|_2 \le \bar{a} \neq 1$, for $i=0,\dots,H-1$, and the disturbance term is bounded as $\|\xi_{t+i}\|_2 \le \gamma$. Then the open-loop prediction error satisfies
\begin{equation}
    \|\Delta_{t+H}\|_2
    \le
    \bar{a}^{H}\|\Delta_t\|_2
    +
    \gamma \sum_{j=0}^{H-1}\bar{a}^{j}
    =
    \bar{a}^{H}\|\Delta_t\|_2
    +
    \gamma \frac{\bar{a}^{H}-1}{\bar{a}-1}.
    \label{eq:open_loop_bound}
\end{equation}
\end{proposition}

Proposition~\ref{prop:growth} gives a geometric upper bound on the open-loop forecasting error. In particular, when $\bar{a}>1$, the bound grows geometrically with the rollout horizon, indicating that long-horizon forecasting can become increasingly sensitive to early prediction errors. This suggests that reducing only one-step prediction error is not sufficient if the autoregressive rollout remains non-contractive.

\subsection{Idealized Closed-Loop Stabilization}
The above result suggests that open-loop forecasting may remain highly sensitive to perturbations over long horizons. A natural way to mitigate this issue is to introduce feedback correction into the rollout process. Consider the idealized closed-loop update
\begin{equation}
    \hat{x}_{t+1}^{\mathrm{cl}} = g_{\theta}(s_t^{\mathrm{cl}}) + \Delta_t^{\mathrm{cl}}.
    \label{eq:ideal_cl_update}
\end{equation}
% Under the same local approximation, the error dynamics become
Substituting the update rule into the error formulation, the closed-loop error dynamics become
\begin{equation}
    \Delta_{t+1}^{\mathrm{cl}} = (A_t - I)\Delta^{\mathrm{cl}}_t + \xi_t.
    \label{eq:ideal_cl_dynamics}
\end{equation}
Compared with the open-loop dynamics in Eq.~(\ref{eq:open_loop_general}), the effective transition operator is shifted from $A_t$ to $A_t-I$. 
See Appendix for a detailed derivation. The stability question is thus transformed into whether feedback can render the local rollout dynamics contractive. The following theorem formalizes the condition under which this closed-loop feedback stabilizes the error dynamics.

\begin{theorem}[Condition for Idealized Closed-Loop Stability]
\label{thm:ideal_stability}
Assume that the disturbance term is uniformly bounded, i.e., $\|\xi_t\|_2 \le \gamma$ for all $t$, and that there exists a constant $q \in [0,1)$ that
\begin{equation}
    \|A_t - I\|_2 \le q, \qquad \forall t.
    \label{eq:contraction_condition}
\end{equation}
Then the idealized closed-loop error satisfies
\begin{equation}
    \|\Delta_{t+H}^{\mathrm{cl}}\|_2
    \le
    q^{H}\|\Delta_t^{\mathrm{cl}}\|_2
    +
    \gamma \frac{1-q^{H}}{1-q},
    \label{eq:ideal_cl_bound}
\end{equation}
and in particular,
\begin{equation}
    \limsup_{H\to\infty}\|\Delta_{t+H}^{\mathrm{cl}}\|_2
    \le
    \frac{\gamma}{1-q}.
    \label{eq:ideal_cl_limsup}
\end{equation}
\end{theorem}

Theorem~\ref{thm:ideal_stability} gives an idealized sufficient condition showing that feedback can prevent unbounded error growth when the corrected local rollout becomes contractive. We stress that this condition is mainly used as a design principle, rather than a direct guarantee for the full nonlinear forecasting system. More analyses and discussion are provided in Appendix.

\subsection{Practical Closed-Loop Forecasting}
The update in Eq.~(\ref{eq:ideal_cl_update}) assumes access to the true residual $\Delta_t$, which is unavailable during inference. In practice, we instead use a residual estimator to predict this residual $\Delta_t$. Let
$\hat{\Delta}_t = \Delta_t + e_t$,
where $e_t$ denotes the estimation error and $\hat{\Delta}_t$ is the output of the residual estimator. Then, the practical closed-loop update is presented as
\begin{equation}
    \hat{x}_{t+1}^{\mathrm{pc}} = g_{\theta}(s_t^{\mathrm{pc}}) + \hat{\Delta}_t^{\mathrm{pc}}.
    \label{eq:practical_cl_update}
\end{equation}
Substituting $\hat{\Delta}_t^{\mathrm{pc}}$ into the error recursion yields
\begin{equation}
    \Delta_{t+1}^{\mathrm{pc}} = (A_t - I)\Delta_t^{\mathrm{pc}} + \xi_t - e_t.
    \label{eq:practical_cl_dynamics}
\end{equation}
Eq.~(\ref{eq:practical_cl_dynamics}) makes that the practical system differs from the idealized one by an additional perturbation term. The following corollary quantifies how this additional estimation error impacts the theoretical upper bound of the long-horizon forecasting performance.

\begin{corollary}[Bounded Error with Imperfect Residual Estimation]
\label{cor:practical_stability}
Assume that the disturbance term satisfies $\|\xi_t\|_2 \le \gamma$, the residual estimation error satisfies $\|e_t\|_2 \le \eta$, and the corrected local rollout satisfies $\|A_t-I\|_2 \le q < 1$ for all $t$. Then the practical closed-loop error satisfies
\begin{equation}
    \|\Delta_{t+H}^{\mathrm{pc}}\|_2
    \le
    q^{H}\|\Delta_t^{\mathrm{pc}}\|_2
    +
    (\gamma + \eta)\frac{1-q^{H}}{1-q},
    \label{eq:practical_cl_bound}
\end{equation}
and consequently,
\begin{equation}
    \limsup_{H\to\infty}\|\Delta_{t+H}^{\mathrm{pc}}\|_2
    \le
    \frac{\gamma + \eta}{1-q}.
    \label{eq:practical_cl_limsup}
\end{equation}
\end{corollary}

Corollary~\ref{cor:practical_stability} extends the idealized closed-loop analysis to a more practical setting by explicitly accounting for residual estimation error. Under the same contractive-feedback condition, the long-horizon forecasting error depends not only on the disturbance level $\gamma$, but also on the residual estimation quality $\eta$. In particular, a smaller estimation error leads to a tighter asymptotic error bound. Therefore, accurate residual prediction is an important factor for making closed-loop correction effective in practice. As with Theorem~\ref{thm:ideal_stability}, this result should be interpreted as an idealized sufficient-condition analysis rather than a literal guarantee for the full nonlinear forecasting system.

\subsection{Implications for the Proposed Framework}\label{4.4}
The theoretical analysis yields two design principles for our framework. First, Eq.~(\ref{eq:practical_cl_limsup}) indicates that the efficacy of closed-loop correction is strictly bounded by the residual estimation error $\eta$. Because the true forecasting error $\Delta_t$ is unobservable during inference, we introduce a \emph{Residual Estimator} to infer the current residual from historical rollouts, thereby providing a real-time feedback signal for online correction.
On the other hand, stable correction necessitates constraining the sensitivity of the rollout dynamics. The contraction condition $\|A_t-I\|_2\leq q<1$ implies the necessary sensitivity range $1-q\leq\|A_t\|_2\leq1+q$; in particular, $\|A_t\|_2\geq2$ is incompatible with contractive direct feedback. Motivated by this observation, we apply Local Lipschitz Regularization to penalize local output amplification and reduce the empirical sensitivity of the predictor, making the rollout dynamics more amenable to feedback correction. More detailed analyses are provided in Appendix.

Therefore, the two components of F-LLM play complementary roles. The Residual Estimator improves the quality of the correction signal, while the Local Lipschitz Regularization reduces the tendency of the rollout process to amplify perturbations. Together, they form the practical design motivated by the above analysis. 
Our analysis is local and based on sufficient conditions. It is intended to explain the source of long-horizon error accumulation and to motivate the proposed design, rather than to provide a global stability guarantee for the full nonlinear forecasting system.

\begin{figure*}
    \centering
    \includegraphics[width=0.85\linewidth]{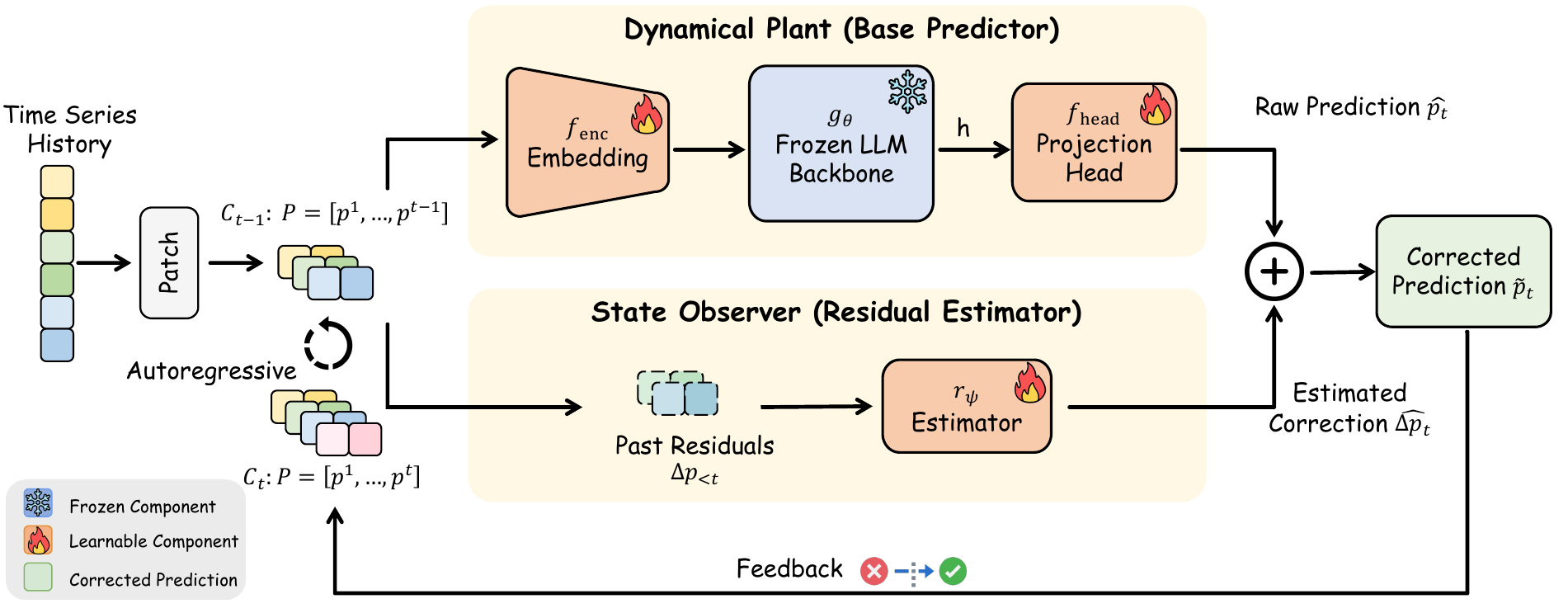}
    \caption{Illustration of F-LLM. The system operates as a closed-loop control mechanism where a frozen LLM generates raw predictions, while a lightweight residual estimator predicts trajectory deviations based on historical errors. The estimated residuals are injected back into the context via an in-loop correction mechanism to mitigate error accumulation.}
    \label{fig:framework}
\end{figure*}

\section{Methodology}\label{sec:method}
Guided by the theoretical constraints derived in Section \ref{sec:theory}, we introduce our framework, \textbf{F-LLM} (Feedback-driven LLM), to mitigate rollout drift by actively modeling and compensating for step-wise generation errors. F-LLM operates as a closed-loop control system composed of three modules: a base predictor acting as the dynamical plant, a residual estimator acting as the state observer, and a feedback mechanism acting as the controller. To explicitly satisfy the theoretical prerequisites for stable correction, we optimize these modules via a tailored two-stage training strategy: first enforcing a local smoothness constraint on the predictor to prevent over-correction, and subsequently training the estimator on this stabilized error distribution to minimize the estimation bound. The framework is provided in Figure \ref{fig:framework}.

\subsection{The F-LLM Framework Design}

\paragraph{Patch-wise Autoregressive Predictor} To align the continuous time series with the semantic space of LLMs, we adopt a patch-based formulation. Given a historical sequence $x \in \mathbb{R}^T$, we segment it into non-overlapping patches $P = [p_1, \dots, p_N]$, where each $p_i \in \mathbb{R}^L$. The base predictor, denoted as the ``Plant'', consists of a learnable embedding layer $f_{\text{enc}}$ to project patches into the LLM's embedding space, a frozen LLM backbone $g_\theta$ to predict future states within this space, and a learnable inverse embedding layer $f_{\text{head}}$ to project the representations back to the continuous time series space. At step $t$, the plant generates a raw prediction based on the autoregressive context window $\mathcal{C}_{t-1}$. This context includes the available information up to step $t-1$, comprising the ground-truth history concatenated with the previously generated (and subsequently corrected) patches. Formally, the raw prediction is:
\begin{equation}
\hat{p}_t = f_{\text{head}}(g_\theta(f_\text{enc}(\mathcal{C}_{t-1}))).
\end{equation}

\paragraph{Residual Estimator} To enact the closed-loop feedback formulated in Section \ref{sec:theory}, the system must compensate for the current prediction error $\Delta p_t$. Naturally, this exact future error is unknown during inference. To bridge this gap, we employ a residual estimator $r_\psi$, implemented as a Multi-Layer Perceptron (MLP), to act as the system observer.

$r_\psi$ infers the required correction from residuals aligned with the current context. At step $t$, the predictor takes a rolling context $\mathcal{C}_{t-1}=[c_{t-N},\ldots,c_{t-1}]$, where each $c_i$ is either an observed historical patch or a previously corrected prediction, and outputs $[\hat{p}_{t-N+1},\ldots,\hat{p}_t]$. The first $N-1$ outputs are aligned with the available context patches $[c_{t-N+1},\ldots,c_{t-1}]$, allowing us to construct the residual history as $\Delta p_{<t}=[c_i-\hat{p}_i]_{i=t-N+1}^{t-1}$. The estimator $r_\psi$ then predicts the correction for the current patch:

\begin{equation}
\hat{\Delta p_t} = r_\psi(\Delta p_{<t}).
\end{equation}
Mathematically, $r_\psi$ parameterizes the theoretical control law derived in Eq.(\ref{eq:practical_cl_update}). Because the true latent state of an autoregressive LLM is partially observable, $r_\psi$ leverages the historical residual sequence $\Delta p_{<t}$ rather than a single step to better infer the system's unobservable drift. By training $r_\psi$ end-to-end, it directly outputs the required compensation term.

\paragraph{Closed-Loop Interaction} The feedback loop is closed by injecting the estimated residual into the prediction. The corrected patch is:
\begin{equation}
\tilde{p}_t = \hat{p}_t + \hat{\Delta p_t}.
\label{eq:in_loop_correction}
\end{equation}
Unlike post-hoc correction methods that only adjust the sequence after generation, F-LLM is an in-loop framework: the corrected patch $\tilde{p}_t$ is appended to the context $\mathcal{C}_t$ for the next autoregressive step. As the context rolls forward, the aligned residual history is recomputed from the updated context at each step. This real-time intervention prevents the ``snowball effect'' by constantly recalibrating the generative trajectory.

\subsection{Two-Stage Training}
Jointly optimizing the base predictor and the residual estimator is unstable due to the ``moving target'' problem: during early training, the randomly initialized $f_\text{enc}$ and $f_{\text{head}}$ produce rapidly shifting, meaningless prediction errors, preventing the estimator from learning stable correction patterns. To resolve this, we adopt a two-stage training strategy. It sequentially fulfills the two theoretical implications discussed in Section \ref{4.4}. In the first stage, we perform Base Predictor Optimization, focusing on bounding the sensitivity of the rollout dynamics. By enforcing the Local Lipschitz constraint, we satisfy the smoothness condition in Theorem \ref{thm:ideal_stability}, thereby constructing a robust dynamical plant that will not over-correct when subjected to direct residual injection. In the second stage, we perform Residual Estimator Training. It freezes the smoothed predictor to yield a stationary error distribution. This allows the optimization to focus on minimizing the residual estimation error $\eta$, ensuring the estimator generates the highly accurate correction signals required for effective closed-loop forecasting.

\paragraph{Stage 1: Base Predictor Optimization} First, we remove the residual estimator and train only the projection layers ($f_{\text{enc}}, f_{\text{head}}$). To stabilize feature alignment, we condition the model strictly on the ground-truth historical sequence rather than its own past predictions. To satisfy the controllability condition in Theorem \ref{thm:ideal_stability} without computationally expensive regularization on the entire LLM, we impose a Local Lipschitz constraint on the actuator $f_{\text{head}}$. Specifically, we perturb the latent output $h$ with a noise vector $\delta$ ($\|\delta\|_2 < \xi$, where $\xi$ denotes the local neighborhood radius) and penalize the amplification of the output difference:
\begin{equation}
\mathcal{L}_\text{lip} 
=
\mathrm{E}_{h, \delta} \left[ \left( \frac{\|f_{\text{head}}(h+\delta) - f_{\text{head}}(h)\|_2}{\|\delta\|_2} - \kappa \right)_+^2 \right],
\end{equation}
where $(\cdot)_+ = \max(0, \cdot)$ penalizes the violations of the Lipschitz bound, $\|\cdot\|_2$ denotes the standard $L_2$ norm, and $\kappa$ is a hyperparameter controlling the desired Lipschitz constant. The objective is the standard reconstruction loss combined with Lipschitz constraint:
\begin{equation}
\mathcal{L}_1 = \frac{1}{N} \sum_{t} \|p_t - \hat{p}_t\|_2 + \lambda \mathcal{L}_{\text{lip}},
\label{eq:stage1_loss}
\end{equation}
where $\lambda$ is a hyperparameter. This stage ensures the base predictor behaves as a robust dynamical system with bounded sensitivity.

\paragraph{Stage 2: Residual Estimator Training} Once the base predictor stabilizes, we freeze its projection layers and activate the residual estimator $r_\psi$. In this stage, $r_\psi$ is trained as a one-step residual predictor over rolling contexts, mapping the context-aligned residual history $\Delta p_{<t}$ to the next residual $\Delta p_t$. The same local correction operator is repeatedly applied during autoregressive inference. By computing the exact step-wise residual $\Delta p_t = p_t - \hat{p}_t$ between the ground truth and the raw prediction, we provide a direct supervisory signal for the estimator. Specifically, we optimize $r_\psi$ to minimize the estimation error:
\begin{equation}
\mathcal{L}_2 = \frac{1}{N} \sum_{t} \| (p_t - \hat{p}_t) - r_\psi(\Delta p_{<t} ) \|_2.
\label{eq:stage2_loss}
\end{equation}
By learning the one-step residual dynamics of the frozen plant, $r_\psi$ produces a feedback signal that is repeatedly injected into the rollout process to counteract autoregressive drift during inference.

To this end, the F-LLM framework instantiates the two design principles derived in Section 4.4. First, the Lipschitz-regularized objective in Eq.~(\ref{eq:stage1_loss}) penalizes local output amplification, improving the predictor's robustness to residual injection. Second, Eq.~(\ref{eq:stage2_loss}) trains the estimator on a stationary error distribution to reduce residual estimation error. Finally, the in-loop feedback mechanism in Eq.~(\ref{eq:in_loop_correction}) continuously recalibrates the autoregressive context. Together, these components reduce error amplification and stabilize long-horizon forecasting in practice.

\section{Experiments}
We conduct extensive experiments to evaluate the effectiveness of F-LLM from multiple perspectives. Specifically, we aim to answer the following questions:
(1) whether in-loop residual feedback improves long-horizon forecasting accuracy under standard supervised settings;
(2) whether such correction preserves the zero-shot generalization ability of frozen LLM forecasters;
and (3) how robust and efficient the proposed framework is across different datasets, LLM backbones, and hyperparameter choices.

\begin{table*}[t]
\centering
\caption{Long-term forecasting results averaged from four prediction lengths $\in$ \{96, 192, 336, 720\}. Detailed results are in Appendix. Most results of other methods derive from \cite{AutoTimes} and \cite{zhang2026enhancing}. Results not reported are obtained by running each method’s source code; the look-back window is selected from \{96, 336, 512, 672\} to report the best results.}
\label{tab_long_res_avg}
\resizebox{1\textwidth}{!}{%
\begin{tabular}{l|cc|cc|cc|cc|cc|cc|cc|cc|cc|cc|cc|cc}
\toprule
\textbf{Dataset} & \multicolumn{2}{c}{\textbf{F-LLM}} & \multicolumn{2}{c}{\textbf{AutoTimes}} & \multicolumn{2}{c}{\textbf{LVICL}} & \multicolumn{2}{c}{\textbf{TimeLLM}} & \multicolumn{2}{c}{\textbf{FPT}} & \multicolumn{2}{c}{\textbf{Unitime}} & \multicolumn{2}{c}{\textbf{iTransformer}} & \multicolumn{2}{c}{\textbf{DLinear}} & \multicolumn{2}{c}{\textbf{PatchTST}} & \multicolumn{2}{c}{\textbf{TimesNet}} & \multicolumn{2}{c}{\textbf{TimeMixer++}} & \multicolumn{2}{c}{\textbf{SimpleTM}} \\
% Year Row
~ & \multicolumn{2}{c}{Ours} & \multicolumn{2}{c}{(2024)} & \multicolumn{2}{c}{(2026)} & \multicolumn{2}{c}{(2024)} & \multicolumn{2}{c}{(2023)} & \multicolumn{2}{c}{(2024)} & \multicolumn{2}{c}{(2023)} & \multicolumn{2}{c}{(2023)} & \multicolumn{2}{c}{(2023)} & \multicolumn{2}{c}{(2023)} & \multicolumn{2}{c}{(2025)} & \multicolumn{2}{c}{(2025)} \\
~ & \textbf{MSE} & \textbf{MAE} & \textbf{MSE} & \textbf{MAE} & \textbf{MSE} & \textbf{MAE} & \textbf{MSE} & \textbf{MAE} & \textbf{MSE} & \textbf{MAE} & \textbf{MSE} & \textbf{MAE} & \textbf{MSE} & \textbf{MAE} & \textbf{MSE} & \textbf{MAE} & \textbf{MSE} & \textbf{MAE} & \textbf{MSE} & \textbf{MAE} & \textbf{MSE} & \textbf{MAE} & \textbf{MSE} & \textbf{MAE} \\
\midrule
\textbf{ETTh1} & \textbf{0.376} & \textbf{0.405} & 0.389 & 0.422 & \underline{0.381} & 0.412 & 0.408 & 0.424 & 0.428 & 0.426 & 0.442 & 0.448 & 0.438 & 0.450 & 0.423 & 0.437 & 0.413 & 0.430 & 0.488 & 0.452 & 0.411 & 0.421 & 0.403 & \underline{0.410} \\
\textbf{ETTh2} & \textbf{0.317} & \textbf{0.367} & 0.352 & 0.395 & \underline{0.326} & 0.376 & 0.354 & 0.393 & 0.353 & 0.391 & 0.356 & 0.399 & 0.382 & 0.414 & 0.431 & 0.447 & 0.330 & 0.379 & 0.414 & 0.427 & 0.333 & \underline{0.371} & 0.338 & 0.374 \\
\textbf{ETTm1} & \textbf{0.319} & \textbf{0.367} & 0.332 & 0.380 & \underline{0.328} & 0.378 & 0.350 & 0.378 & 0.366 & 0.382 & 0.375 & 0.403 & 0.370 & 0.399 & 0.357 & 0.379 & 0.351 & 0.381 & 0.400 & 0.406 & 0.360 & \underline{0.369} & 0.363 & 0.378 \\
\textbf{ETTm2} & \textbf{0.234} & \textbf{0.300} & 0.243 & 0.310 & \underline{0.239} & \underline{0.306} & 0.254 & 0.314 & 0.265 & 0.315 & 0.277 & 0.329 & 0.272 & 0.331 & 0.267 & 0.334 & 0.255 & 0.315 & 0.291 & 0.333 & 0.264 & 0.313 & 0.264 & 0.308 \\
\textbf{ECL} & \textbf{0.148} & \textbf{0.240} & 0.159 & 0.253 & \underline{0.158} & \underline{0.248} & 0.159 & 0.253 & 0.167 & 0.263 & 0.216 & 0.304 & 0.177 & 0.268 & 0.177 & 0.274 & 0.159 & 0.253 & 0.193 & 0.295 & 0.161 & \underline{0.248} & \underline{0.158} & \underline{0.248} \\
\textbf{Weather} & \textbf{0.215} & \textbf{0.251} & 0.235 & 0.273 & 0.222 & 0.262 & 0.225 & {0.258} & 0.236 & 0.271 & 0.248 & 0.275 & 0.238 & 0.273 & 0.240 & 0.300 & 0.225 & 0.264 & 0.259 & 0.286 & \underline{0.220} & \underline{0.256} & 0.236 & 0.265 \\
\textbf{Traffic} & \textbf{0.364} & \underline{0.261} & 0.374 & 0.264 & \underline{0.370} & \textbf{0.260} & 0.388 & 0.264 & 0.414 & 0.294 & 0.429 & 0.304 & 0.379 & 0.272 & 0.434 & 0.295 & 0.391 & 0.264 & 0.620 & 0.329 & 0.409 & \textbf{0.260} & 0.422 & 0.276 \\
\bottomrule
\end{tabular}
}
\end{table*}

\subsection{Time Series Forecasting}

\paragraph{Experimental Setup.}
We evaluate F-LLM on seven real-world multivariate time series datasets from the electricity, weather, and economy domains (Details in Appendix). Hyperparameters such as learning rate, batch size, and hidden dimension are selected from pre-defined grids based on performance. We comprehensively compare F-LLM with representative forecasting models under standard settings in \cite{AutoTimes}, where the lookback length is fixed to 672 and prediction horizons are $\{96, 192, 336, 720\}$. 
Baselines include conventional deep learning baselines (PatchTST \cite{PatchTST}, DLinear \cite{DLinear}, TimesNet \cite{TimesNet}, SimpleTM \cite{chen2025simpletm}, TimeMixer++ \cite{wang2025timemixer++} and iTransformer \cite{iTransformer}) to recent LLM-based approaches (TIME-LLM \cite{Time-LLM}, AutoTimes \cite{AutoTimes}, UniTime \cite{UniTime}, LVICL \cite{zhang2026enhancing}, and FPT \cite{FPT}).

\paragraph{Comparison Results.}
As Table~\ref{tab_long_res_avg} shows, across the seven datasets, F-LLM achieves the best performance in most horizons. Compared with the strongest baseline on each setting (including both LLM4TS and non-LLM models), it has lower MSE on the most challenging long-horizon cases. Notably, these gains are obtained by adding a lightweight feedback module on top of frozen LLM forecasters, without modifying or retraining the backbone models.

\subsection{Zero-Shot Forecasting}
Large language models exhibit remarkable zero-shot generalization~\cite{gpt3}. To assess whether this property transfers to TSF, we evaluate F-LLM in a zero-shot setting, where no samples from the target domain are used during training. Following the benchmark protocol of AutoTimes~\cite{AutoTimes}, a model is trained on a source dataset and applied to an unseen target domain without fine-tuning. 

\paragraph{Experimental Setup.}
We conduct transfer learning between the M3 and M4 datasets, both rich in temporal variation but differing in statistical distribution. Each experiment includes a source–target pair (e.g., M4~Monthly $\to$~M3~Monthly). For M4~$\to$~M3, models trained on M4~Yearly, Quarterly, and Monthly subsets are evaluated on their M3 counterparts, while M3~Others uses the M4~Quarterly model to align forecast horizons. For M3~$\to$~M4, the process is symmetric: M4~Yearly, Quarterly, and Monthly subsets use corresponding M3 models, and the remaining subsets (M4~Weekly, Daily, Hourly) adopt the model trained on M3~Monthly.

\paragraph{Comparison Results.}
As shown in Table~\ref{tab:forecast_zeroshot_full}, F-LLM attains the lowest SMAPE on 7 out of 8 transfer settings and is second best on the remaining one (M4$\to$M3 Others, where FPT is slightly better). It consistently outperforms or matches both LLM-based forecasters (AutoTimes, FPT) and strong deep TSF baselines (e.g., DLinear and Transformer variants), indicating that the proposed bias correction module preserves the zero-shot generalization ability of LLMs while further improving cross-domain forecasting accuracy.

% \subsection{Ablation Study}
% 1.有校正 vs 无校正的结果对比以及可视化对比
% 2.不同backbone比较
% 2.1 去掉LLM比较
% 3.不同patch长度的精度对比
% 4.不同历史长度的结果对比
% 5.运行时间对比，优化参数对比，鲁棒性对比

% 待改1：增加与Post-hoc Residual Correction、Fine-tuning（可以只调LoRA / Adapter或只调 embedding）的对比

% 待改2：增加Residual autocorrelation可预测实验：将Autotimes的预测结果可视化，展示residual 是否存在“滞后相关”，
% 白噪声的理论特征：除 lag=0 外，自相关应接近 0。证明自回归残差并不是白噪声
% 若多个 lag 明显超出置信区间（比如 95%），则 residual/bias 不是白噪声 → 有结构可利用。

% 待改3：看 residual 是否存在周期性或频带能量集中。白噪声在频域上应接近“平坦谱”（flat spectrum）。
% 若存在明显谱峰（例如日周期、周周期等）或者能量集中在某些频段⇒ residual 不是白噪声。

% 待改4：“简单模型 vs F-LLM 的 bias predictor”：只要 residual 可预测，那么一个很简单的模型也应当能比“零预测”更好。
% 增加对比：（1）Zero / Mean baseline、（2）Persistence baseline（沿用上一个）、（3）Linear baseline用AR(1)/AR(p) 或 Ridge regression 与F-LLM对比
% 评估指标：（1）bias prediction 的 MAE/MSE：$|b-\hat{b}|$ （2）以及最终 forecast 的 MAE/MSE（加上 correction 后的效果）
\subsection{Ablation and Extended Analysis}
We further conduct a series of ablation and sensitivity experiments to analyze the effectiveness, and efficiency.

\begin{figure}
    \centering
    \subfloat[ETTh1]{
        \includegraphics[width=0.48\linewidth]{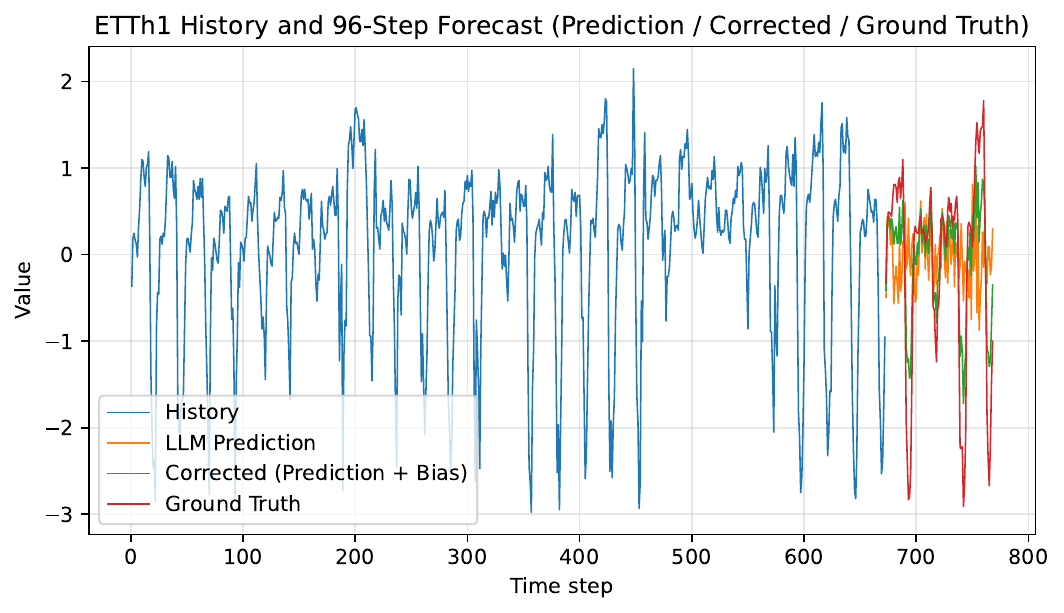}
    }
    \subfloat[Weather]{
        \includegraphics[width=0.48\linewidth]{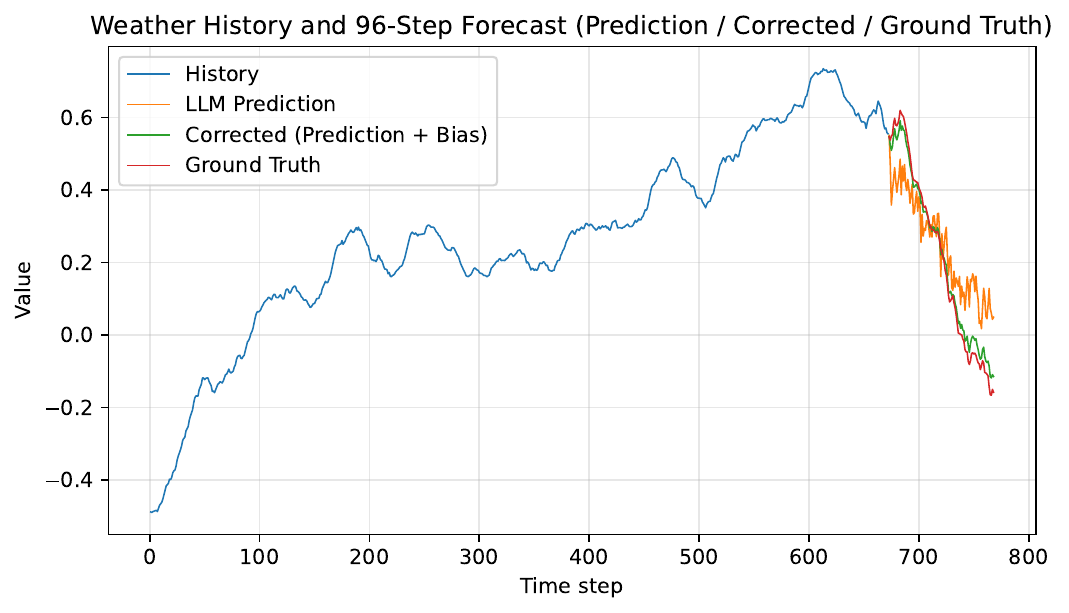}
    }
    \caption{Comparison of 96-step forecasts on ETTh1 and Weather. F-LLM compensates for structured residual bias and maintains trajectory alignment over long horizons.}
    \label{fig:bias_vis}
\end{figure}

\paragraph{Effectiveness of each module.}
To assess the contribution of each component in F-LLM, we compare three variants: (1) F-LLM w/o Feedback: removing the bias correction module, (2) F-LLM w/o LLM: replacing the LLM forecaster with an MLP, and (3) F-LLM (Full): the complete model. As shown in Table~\ref{tab:abla_bc}, eliminating either component consistently degrades performance across all horizons on both ETTh1 and Weather. We further compare F-LLM with post-hoc multi-step residual correction and a variant without Local Lipschitz Regularization in Appendix. Both variants are consistently weaker than F-LLM across all horizons, supporting the complementary roles of in-loop feedback and sensitivity control. Figure~\ref{fig:bias_vis} further visualizes 96-step forecasts, where the uncorrected LLM output exhibits gradual drift from the ground truth, whereas F-LLM adaptively offsets these deviations.

\begin{table*}[t]
  % \vspace{5pt}
  \centering
    \caption{Results of zero-shot forecasting. We adopt the same protocol as FPT. M4 $\to$ M3 means training forecasters on M4 datasets and evaluating the performance on M3, and vice versa. Lower SMAPE indicates better performance.}
  \label{tab:forecast_zeroshot_full}
  \renewcommand{\multirowsetup}{\centering}
  \resizebox{0.75\textwidth}{!}{
  \begin{tabular}{c|c|cccccccccc}
    \toprule
    \multicolumn{2}{c|}{Method} & \textbf{F-LLM} & {AutoTimes} & FPT & DLinear & PatchTST & TimesNet & NSformer & FEDformer & Informer & Reformer  \\ 
    \toprule
    \multirow{4}{*}{\rotatebox{90}{M4$\ \to\ $M3}} 
    & Yearly  & \textbf{15.08}  & \underline{15.71} & 16.42 & 17.43 & {15.99} & 18.75 & 17.05 & 16.00 & 19.70 & 16.03\\
    & Quarterly & \textbf{9.06} & \underline{9.35} & 10.13 & 9.74 & 9.62 & 12.26 & 12.56 & {9.48} & 13.00 & 9.76 \\
    & Monthly & \textbf{13.68}  & {14.06} & 14.10 & 15.65 & 14.71 & \underline{14.01} & 16.82 & 15.12 & 15.91 & 14.80 \\
    & Others & \underline{5.08} & {5.79} & \textbf{4.81} & 6.81 & 9.44 & 6.88 & 8.13 & 8.94 & 13.03 & 7.53 \\
    % & Average & \textbf{10.90} & \underline{12.75} & {13.06} & 14.03 & 13.39 & 14.17 & 15.29 & 13.53 & 15.82 & 13.37 \\
    \midrule
    \multirow{4}{*}{\rotatebox{90}{M3$\ \to\ $M4}} 
    & Yearly    & \textbf{13.701} & \underline{13.728} & {13.740} & 14.193 & 13.966 & 15.655 & 14.988 & 13.887 & 18.542 & 15.652\\
    & Quarterly & \textbf{10.723} & \underline{10.742} & {10.787} & 18.856 & 10.929 & 11.877 & 11.686 & 11.513 & 16.907 & 11.051\\
    & Monthly   & \textbf{14.535} & \underline{14.558} & {14.630} & 14.765 & 14.664 & 16.165 & 16.098 & 18.154 & 23.454 & 15.604 \\
    & Others  & \textbf{6.215} & \underline{6.259} & 7.081 & 9.194 & 7.087 & {6.863} & 6.977 & 7.529 & 7.348 & 7.001\\
    % & Average & \textbf{0} & \underline{13.036} & {13.125} & 15.337 & 13.228 & 14.553 & 14.327 & 15.047 & 19.047 & 14.092\\
    \bottomrule
  \end{tabular}}
\end{table*}

\begin{table}[t]
  \centering
    \caption{Ablation study on the effectiveness of individual components in F-LLM on the ETTh1 and Weather datasets.}\label{tab:abla_bc}
  \renewcommand{\multirowsetup}{\centering}
  \resizebox{0.4\textwidth}{!}{\begin{tabular}{c|c|cc|cc}
    \toprule
    \multicolumn{2}{c|}{Dataset} & 
    \multicolumn{2}{c}{\rotatebox{0}{\scalebox{1.0}{ETTh1}}} &
    \multicolumn{2}{c}{\rotatebox{0}{\scalebox{1.0}{Weather}}}\\
     \cmidrule(lr){1-2} \cmidrule(lr){3-4} \cmidrule(lr){5-6}
    \multicolumn{2}{c|}{Metric} & \scalebox{1.0}{MSE} & \scalebox{1.0}{MAE}  & \scalebox{1.0}{MSE} & \scalebox{1.0}{MAE}  \\
    \toprule
    \multirow{3}{*}{Pred-$96$} 
    & F-LLM w/o Feedback & {0.362} & 0.395 & 0.147 & 0.204 \\
    & F-LLM w/o LLM & {0.377} & 0.406 & 0.149 & 0.215 \\
    & \textbf{F-LLM (Full)} & \textbf{0.347} & \textbf{0.389} & \textbf{0.140} & \textbf{0.188} \\
    \midrule
    \multirow{2}{*}{Pred-$192$} 
    & F-LLM w/o Feedback & {0.390} & {0.415} & 0.203 & 0.254\\
    & F-LLM w/o LLM & {0.405} & {0.418} & 0.216 & 0.268\\
    & \textbf{F-LLM (Full)} & \textbf{0.375} & \textbf{0.402} & \textbf{0.185} & \textbf{0.231}\\
    \midrule
    \multirow{2}{*}{Pred-$336$} 
    & F-LLM w/o Feedback & {0.405} & {0.434} & 0.245 & 0.306\\
    & F-LLM w/o LLM & {0.416} & {0.443} & 0.255 & 0.311\\
    & \textbf{F-LLM (Full)} & \textbf{0.390} & \textbf{0.415} & \textbf{0.230} & \textbf{0.270}\\
    \midrule
    \multirow{2}{*}{Pred-$720$} 
    & F-LLM w/o Feedback & 0.417 & 0.444 & 0.320 & 0.341  \\
    & F-LLM w/o LLM & 0.438 & 0.460 & 0.324 & 0.351 \\
    & \textbf{F-LLM (Full)} & \textbf{0.395} & \textbf{0.427} & \textbf{0.305} & \textbf{0.318}\\
    \bottomrule
  \end{tabular}}
% \vspace{-10pt}
\end{table}

\begin{figure}
    \centering
    \includegraphics[width=1.0\linewidth]{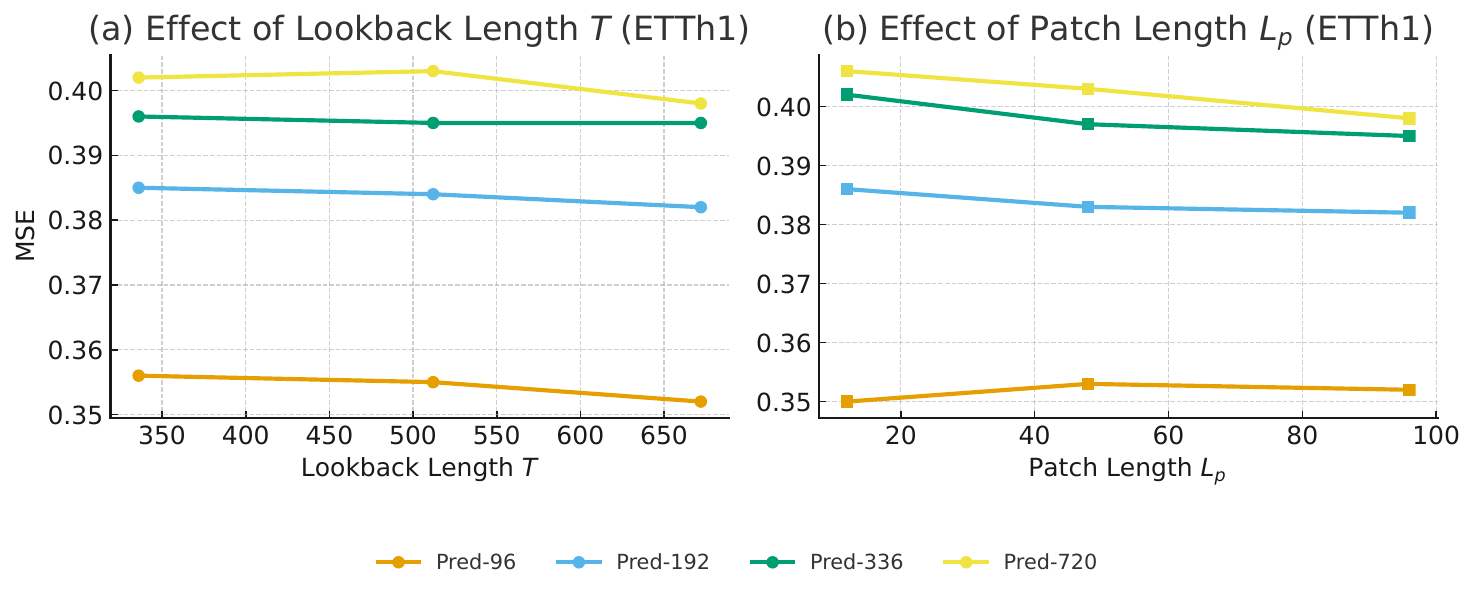}
    \caption{Sensitivity analysis of F-LLM to the historical lookback length $T$ and patch length $L_p$ on the ETTh1 dataset.}
    \label{fig:hyper_length}
\end{figure}

\paragraph{Hyperparameter Sensitivity.}
We further investigate the sensitivity of F-LLM to two critical hyperparameters that influence forecasting accuracy: the historical lookback length $T$ and the patch length $L_p$. Both parameters jointly determine the effective temporal range and tokenization granularity of the LLM-based forecaster. Specifically, we vary the lookback length $T \in \{336, 512, 672\}$ and the patch length $L_p \in \{12, 48, 96\}$ on the ETTh1 dataset, and report the corresponding MSE across four forecasting horizons. The results are summarized in Figure~\ref{fig:hyper_length}. As $T$ increases, the model performance gradually improves and then stabilizes, indicating that extending the historical context benefits prediction up to a certain point. Regarding the patch length, moderate segment sizes (around $L_p=48$) yield the most favorable results, while overly short or long patches slightly reduce accuracy. These observations suggest that F-LLM maintains stable performance under a range of hyperparameter choices, and that the patching design is not the main source of gains but mainly helps capture local temporal patterns before feeding them to the frozen LLM backbone.

% \begin{table}
% \centering
% \resizebox{0.9\linewidth}{!}{
% \begin{tabular}{llccc}
% \toprule
% Backbone Model & Train Time (s/iter) & Test Time (s/iter) & Tunable Parameters(M) \\
% \midrule
% (LLaMA-7B) &
% F-LLM & 0.437 & 0.211 & 1.05 \\
% & Autotimes & 0.354 & 0.163 & 0.79 \\
% & TimeLLM & 1.896 & 0.734 & 45.66 \\
% \bottomrule
% \end{tabular}}
% \caption{Efficiency comparison between F-LLM and representative LLM4TS methods in terms of training and inference time as well as the number of tunable parameters, measured with the same batch size (224) on the ETTh1 dataset. F-LLM introduces only marginal computational overhead by adding a lightweight inference-time feedback module on top of frozen LLM backbones.}
% \label{tab:efficiency}
% %\vspace{-10pt}
% \end{table}

% \begin{figure}
%     \centering
%     \includegraphics[width=1.0\linewidth]{Fig/efficiency_comparison.pdf}
%     \caption{Efficiency comparison between F-LLM and representative LLM4TS methods in terms of training and inference time as well as the number of tunable parameters, measured with the same batch size (224) on the ETTh1 dataset.}
%     \label{fig:efficiency}
% \end{figure}

\begin{figure}
    \centering
    \subfloat[Inference Time]{
        \includegraphics[width=0.43\linewidth]{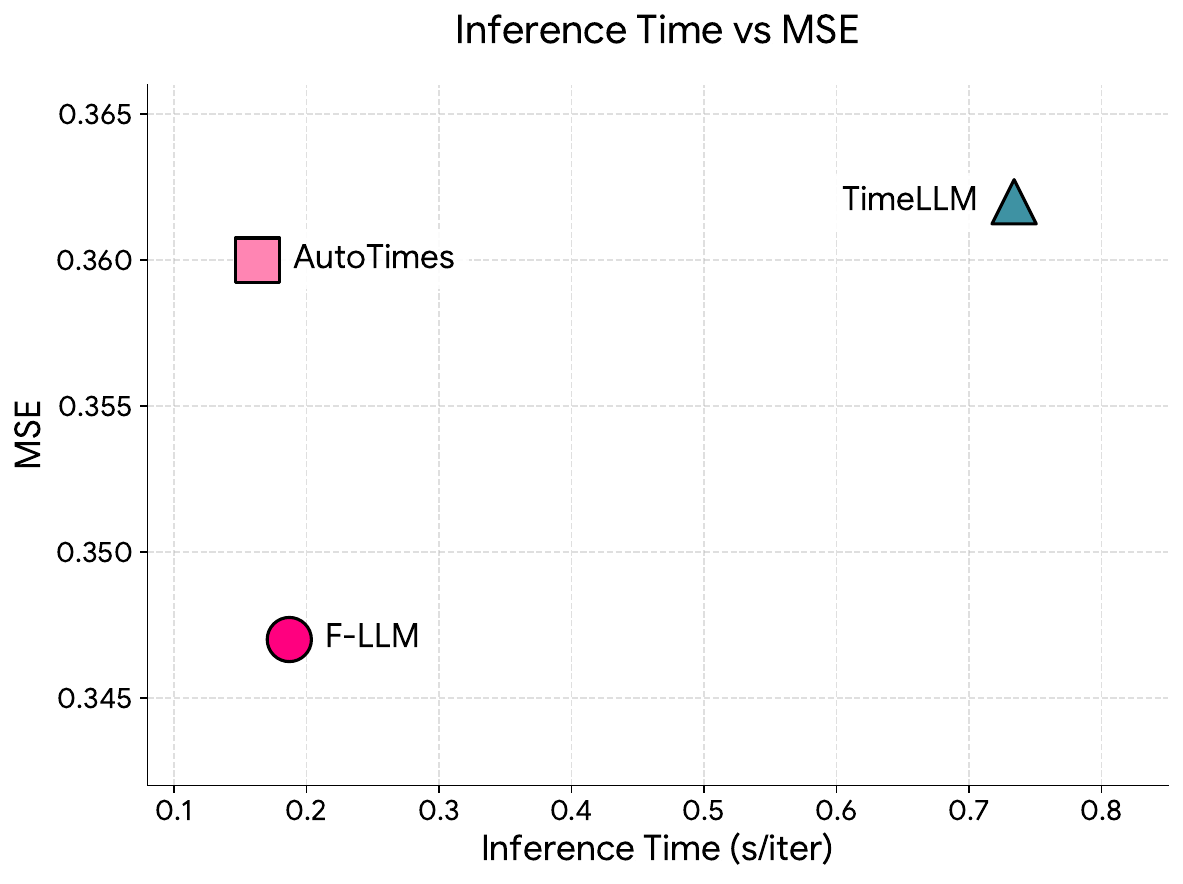}
    }
    \subfloat[Tunable Parameters]{
        \includegraphics[width=0.43\linewidth]{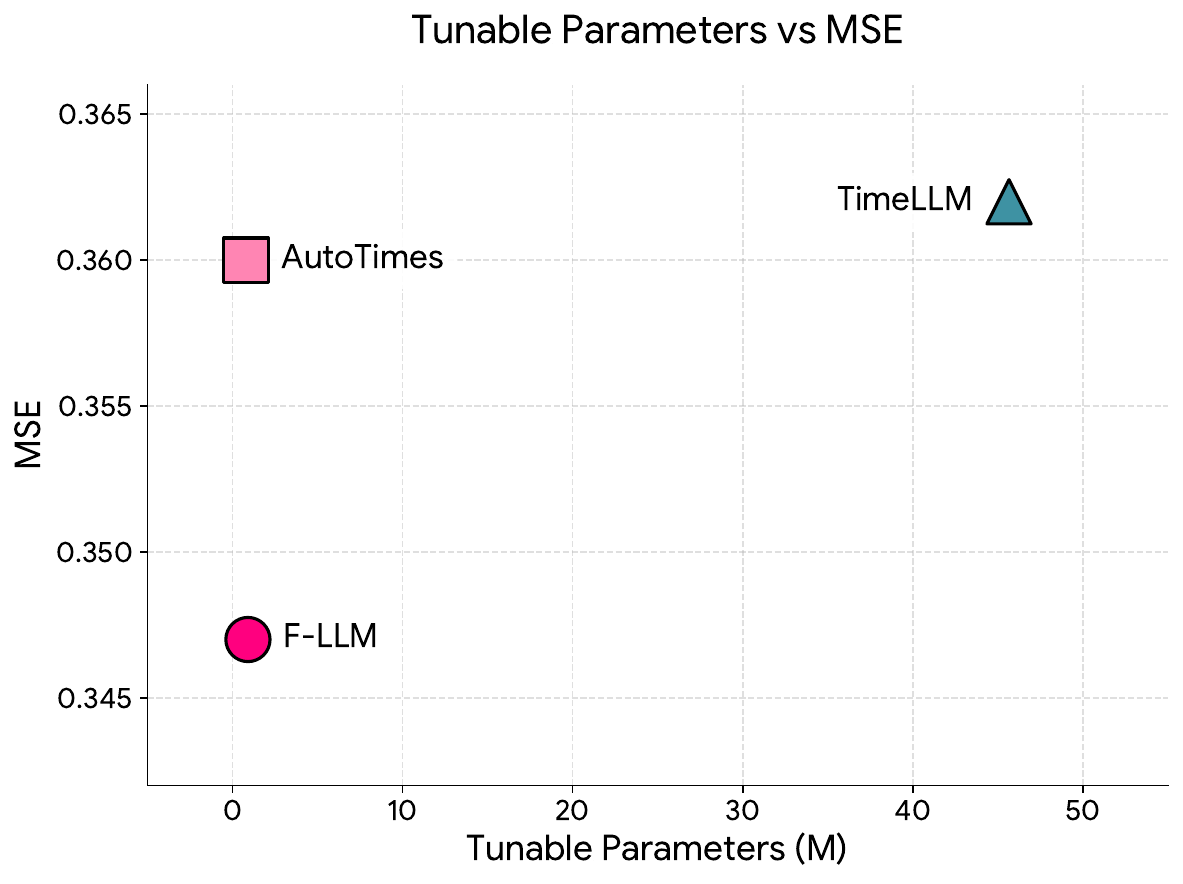}
    }
    \caption{Efficiency comparison between F-LLM and representative LLM4TS methods in terms of inference time as well as the number of tunable parameters on the ETTh1 dataset.}
    \label{fig:efficiency}
\end{figure}

\paragraph{Efficiency Analysis.}
We further evaluate the efficiency and robustness of F-LLM compared with representative LLM4TS baselines. As shown in Figure~\ref{fig:efficiency}, F-LLM consistently improves forecasting accuracy with negligible computational cost. Furthermore,  to disentangle the effect of inference-time residual correction from parameter-efficient fine-tuning, we conduct a controlled comparison among three settings: (1) the frozen LLM baseline, (2) the frozen LLM baseline equipped with LoRA, and (3) F-LLM with residual feedback correction. As shown in Table~\ref{tab:correction_lora}, F-LLM consistently achieves larger gains, especially in long-horizon forecasting. These results indicate that correcting structured prediction bias at inference time is more effective than lightweight parameter adaptation for mitigating error accumulation in autoregressive forecasting.

\begin{table}[t]
\centering
\caption{Comparison of long-term forecasting performance under different adaptation strategies.}
\label{tab:correction_lora}
\resizebox{0.9\linewidth}{!}{
\begin{tabular}{c|c|cc|cc|cc}
\toprule
Dataset & Metric &
\multicolumn{2}{c}{ETTh1} &
\multicolumn{2}{c}{ECL} &
\multicolumn{2}{c}{Weather} \\
\cmidrule(lr){3-4} \cmidrule(lr){5-6} \cmidrule(lr){7-8}
& & MSE & MAE & MSE & MAE & MSE & MAE\\
\midrule
\multirow{3}{*}{Pred-96}
& Frozen LLM      & 0.362 & 0.395 & 0.142 & 0.235 & 0.147 & 0.204\\
& + LoRA        & 0.358 & 0.395 & 0.131 & 0.221 & 0.145 & 0.195\\
& F-LLM & \textbf{0.347} & \textbf{0.389} & \textbf{0.117} & \textbf{0.204}
                          & \textbf{0.140} & \textbf{0.188}\\
\midrule
\multirow{3}{*}{Pred-192}
& Frozen LLM      & 0.390 & 0.415 & 0.161 & 0.251 & 0.203 & 0.254\\
& + LoRA        & 0.390 & 0.416 & 0.147 & 0.244 & 0.198 & 0.242\\
& F-LLM & \textbf{0.375} & \textbf{0.402} & \textbf{0.139} & \textbf{0.230}
                          & \textbf{0.185} & \textbf{0.231} \\
\midrule
\multirow{3}{*}{Pred-336}
& Frozen LLM      & 0.405 & 0.434 & 0.174 & 0.269 & 0.245 & 0.306\\
& + LoRA        & 0.406 & 0.434 & 0.168 & 0.255 & 0.238 & 0.291\\
& F-LLM & \textbf{0.390} & \textbf{0.415} & \textbf{0.150} & \textbf{0.248}
                          & \textbf{0.230} & \textbf{0.270} \\
\midrule
\multirow{3}{*}{Pred-720}
& Frozen LLM      & 0.417 & 0.444 & 0.218 & 0.302 & 0.320 & 0.340\\
& + LoRA        & 0.414 & 0.442 & 0.207 & 0.290 & 0.311 & 0.325\\
& F-LLM & \textbf{0.395} & \textbf{0.427} & \textbf{0.186} & \textbf{0.279}
                          & \textbf{0.305} & \textbf{0.318}\\
\bottomrule
\end{tabular}
}
\end{table}

\section{Conclusion}
In this paper, we identify the structural vulnerability of error accumulation inherent in applying autoregressive LLMs to continuous time series forecasting. To mitigate the rollout drift caused by open-loop inference, we propose F-LLM, a novel framework that reformulates forecasting as a closed-loop control problem. By integrating a residual estimator as a system observer and imposing Local Lipschitz constraints to promote controllability, F-LLM effectively rejects disturbances and actively calibrates the forecasting trajectory. Our theoretical analysis proves that, under explicit contraction assumptions, this closed-loop mechanism guarantees a uniformly bounded step-wise error sequence within the local surrogate dynamics. Extensive experiments conducted on various benchmark datasets confirm that F-LLM achieves superior performance on standard benchmarks while successfully retaining the zero-shot generalization capabilities of the frozen LLM backbone.

% Extensive experiments confirm that F-LLM not only achieves state-of-the-art performance on long-horizon benchmarks but also preserves the zero-shot generalization capabilities of frozen LLMs with minimal computational overhead.

\begin{acks}
The authors sincerely thank the anonymous reviewers for their valuable comments and constructive feedback. This work was supported by the National Natural Science Foundation of China under Grant No.~62506355 and Grant No.~62501581.
\end{acks}
%%
%% The next two lines define the bibliography style to be used, and
%% the bibliography file.

\bibliographystyle{ACM-Reference-Format}
\bibliography{reference}

\end{document}